\documentclass[letterpaper, 10 pt, conference]{ieeeconf}  

\IEEEoverridecommandlockouts                              

\usepackage{amsmath}
\usepackage{amssymb}
\usepackage{booktabs}

\usepackage{url}

\usepackage{tikz}

\newcommand\copyrighttext{%
  \footnotesize \textcopyright 2026 IEEE. Personal use of this material is permitted.
  Permission from IEEE must be obtained for all other uses, in any current or future
  media, including reprinting/republishing this material for advertising or promotional
  purposes, creating new collective works, for resale or redistribution to servers or
  lists, or reuse of any copyrighted component of this work in other works.}
\newcommand\copyrightnotice{%
\begin{tikzpicture}[remember picture,overlay]
\node[anchor=south,yshift=10pt] at (current page.south) 
  {\fbox{\parbox{\dimexpr\textwidth-\fboxsep-\fboxrule\relax}{\copyrighttext}}};
\end{tikzpicture}%
}

\title{\LARGE \bf
Stop Pretending Social Robots Are Inevitable
}

\author{Serge Thill\\Donders Institute for Brain, Cognition, and Behaviour\\Radboud University Nijmegen\\6525GD Nijmegen, The Netherlands\\
        {\tt\small serge.thill@donders.ru.nl}}

\begin{document}
\maketitle
\copyrightnotice

\begin{abstract}
This paper takes issue with the recent themes of both the RO-MAN and the HRI conferences for their portrayal of a future human-robot society as inevitable. The focus is on discussing how such statements ultimately shape research. By treating a future human-robot society as a fait accompli, license is given for user studies to imagine any scenario they like, no matter whether it has any ecological relevance, and to emphasise the scenario design over actually creating robot abilities needed to fullfill the imagined role. Meanwhile, research that focusses on actual societal needs, without assuming that robots are a solution, is deprioritised, as is technical development, in particular with respect to abilities that are necessary to enable robots that function as social agents rather than a mere automation of tasks. A frame that simply assumes a robot future not only detracts from scientific advancement in favour of a techno-solutionism we ought to resist, it is also self-defeating as it risks stifling the research needed to bring it about. We should therefore reject attempts to frame and promote the field in terms of the inevitable social robot and instead focus on one that facilitates advances in the field regardless of what the future holds. This paper suggests that a renewed focus on cognitive mechanisms necessary for the ``I'' in HRI would be a good starting point.
\end{abstract}

\section{Introduction}

We need to talk about the assumption that a future human-robot society is inevitable. The theme of this very conference is \emph{“Realizing Human-Robot Symbiosis with AI.”} and the accompanying blurb\footnote{Retrieved on February 5 from\\\url{https://ro-man2026.org/call-for-papers/}} reads (emphases added):

\begin{quote}
As robots become increasingly capable of socially interacting with people, it is important to think about how we can shape a future in which humans and robots collaborate and work together as teammates. Rather than replacing us, our vision is that \textbf{robots will complement us}. We envision a hybrid intelligence where robots and humans together can be more capable, effective, productive, efficient, and enjoyable to work with. To realize a successful integration of socially interactive robots into our society, \textbf{we need to invest in a future where human–robot teams become prevalent}. To this end, we aim to create a clear vision of this future and foster innovations in interaction models, social capabilities, and robot skills that \textbf{help robots understand how they can work together with us}.
\end{quote}

Note the assertions (in bold) that assume a need for robots rather than ask if there is one. In the field more broadly, this is not a one-off. Another HRI (the field) conference, HRI (the conference), had a very similar theme, likewise simply  assuming the societal benefit of robots (``Empowering Society'') and a statement\footnote{Retrieved on February 5 from\\\url{https://humanrobotinteraction.org/2026/}} centred on how to best convince society to accept the field's techno-solutionist view:

\begin{quote}
Our field has the potential to bring about positive change in many areas of our societies such as healthcare, transport, remote working, agriculture and industry. However, this change cannot happen if we do not engage properly with the end users who will potentially utilize robots in their jobs and daily lives.

For this reason, HRI 2026 will focus on: 1) \textbf{how we can ethically integrate robots in everyday processes} without creating disruptions or inequalities, carefully thinking at the future of work and services; 2) \textbf{how we can make them accessible to the general public} (in terms of design, technical literacy and cost) \textbf{with the final aim to make robots more willingly adopted as technological helpers}.
\end{quote}

The HRI conference first used a theme that stated the inevitability of social robots (and assumed benefits to society) in 2010 (``Robots are becoming part of people’s everyday social lives — and will increasingly become so.''\footnote{See \url{https://humanrobotinteraction.org/hri-2010/}}). For RO-MAN, it is harder to trace back older themes, but in recent years, the inevitable social robot was a constant presence at both HRI and RO-MAN. At HRI, it was visible in 2025 (``Robots for a Sustainable World''), 2024 (``HRI in the real world) and 2023 (``HRI for all''). RO-MAN had a similar message in 2025 (``Shaping our hybrid future with robots together''), and although 2024 (``Embracing Human-Centered HRI'') still emphasised the ``need to understand human needs'', it was also with a view towards a collective effort that is `` focused on enhancing human quality of life, now and for future generations'''\footnote{Quotes from the Call for Papers at the time; the conference website appears to no longer be available}.

The assumption that social robots are somehow inevitable is not limited to conference themes; it also increasingly appears in introductory sections of papers, as has been pointed out several times before \cite{Gamboa2025,Zawieska2024}). Critically, when others have pointed this out, like Gamboa or Zawieska in the papers just cited, they did so because there is nothing to justify this assumption. And yet, it somehow perseveres. Hence this paper.

The main point in the remainder is to demonstrate not just that the assumption that social robots are inevitable is not justified (which, again, is not a new observation), but that this frame risks being actively harmful to HRI as a research field because it encourages papers that do little to actually advance the field while stifling activities that would. HRI, as a research field, is not without its problems \cite{Gamboa2025,Zawieska2024,LemaignanEtAl2026,MatheusScassellati2026,BethelMurphy2010,LeichtmannEtAl2022,BartlettEdmundsBelpaeme2022}, many of which, as we will discuss, can follow from assertions that these robots will definitely exist. An old analogy \cite{Forscher1963} for science likens \emph{facts} to bricks and \emph{explanations} to edifices, lamenting (in 1963!) both the proliferation of brick production at the expense of edifices, and the confusion of piles of bricks for, again, edifices. Claiming social robots are inevitable lets us produce lots of bricks that we can place in piles, but how these should come together to form an edifice seems increasingly unclear. As such, this paper is also about asking what we, as a research community, would like the field of HRI to be about.

\section{From Conference Theme to Research}

Any scientific field is shaped by assumptions. In formulating a theory for a field, being clear about such assumptions is a necessary step \cite{Guest2024}, and there is value in exploring the consequences of such assumptions \cite{McClelland2009}. When it comes to HRI, the first issue is that empirical work often seems to lack an underlying theory \cite{LeichtmannEtAl2022}. What appears to be guiding instead, given the opening sentences of papers \cite{Gamboa2025,Zawieska2024}, is the assumption that the world is inevitably moving towards a future in which social robots are common in society.

There is clearly a space for visions of future society in research. If nothing else, they help to determine funding priorities \cite{BorupEtAl2006}. Thinking about the future is, however, a science in itself \cite{Gabriel2013,LeeEtAl2021} and not every imaginable future is a realistic one. Where exactly the belief that social robots are inevitable comes from is difficult to trace back other than noting the afore-mentioned conference themes and increasing statements to this effect in papers \cite{Gamboa2025,Zawieska2024}. A scientifically motivated reason for this claim seems to be lacking. It may well be that a claim once used to convince funding bodies to invest in HRI turned into something that now reassures the field itself of its relevance.

Whatever the origin, a frame with no clear connection to theory now shapes the field. Conference themes signal what kind of paper is welcome; funding priorities directly impact the proposed research. The perceived lack of theory in HRI is in itself a concern since good theories are necessary, but no simple matter \cite{Guest2024,GuestMartin2021a}. The impact of a badly chosen frame is, however, more general: visions lead to hypotheses, and bad hypotheses can be a liability \cite{FelinEtAl2021}, just like faulty, or badly assembled bricks cause an edifice to crumble \cite{Forscher1963}. An inadequate vision can set the field up for research that does not lead to progress or is quickly obsolete \cite{BigmanEtAl2026}, turning hype into disappointment \cite{BorupEtAl2006}. Smaller research communities, and HRI is arguably one, are more susceptible to this \cite{BorupEtAl2006}. As an example of issues that can arise, consider research in embodied cognitive science, which, two decades ago or so, set itself up to encourage studies that show involvement of (pre)motor brain areas in higher cognition. On the one hand, this led to a proliferation of studies to show just that. On the other, because of the way the hypothesis space was cast, those did little to advance the field \cite{MahonCaramazza2008}. By declaring social robots inevitable at the outset, HRI sets itself up for a similar situation: on the one hand, there will always be another study to run, another brick to make \cite{Forscher1963}; on the other, it is not clear how exactly it might advance the field. In the following, we explore some of the pitfalls our framing enables.

\section{How The Inevitable Social Robot Shapes Work in HRI}

\subsection{Meaningful participatory research suffers}

Most immediately, treating social robots as inevitable deprioritises research into what society actually needs in favour of techno-centric solutionism. By assuming a robot future, the field also makes decisions about the future for, rather than with, society. This is particularly evident in the conference themes quoted at the outset that only ask how to make society accept a social robot, not whether it even wants one. When stakeholders do get involved, it is thus typically only to investigate how exactly to introduce this robot, or what the preferred robot would be. User studies in this sense may occasionally involve the public, but they are not meaningful examples of stakeholder involvement and the transdisciplinary research this entails \cite{vanderBijl-Brouwer2022}. The HRI community is a very specific subset of society as a whole, with ties to specific industries (discussed more below), and we need to ask ourselves, as some do \cite{Gamboa2025}, why we should be the ones to decide that robots are beneficial, and a human-robot future inevitable and desirable. In other fields at the intersection of technology and society, the problematic nature of a techno-centric field (or industry) preaching solutions already receives more attention. This is particularly salient in AI \cite{birhaneValuesEncodedMachine2022} and specific application domains thereof,  such as education \cite{Selwyn2022,PeetersEtAl}. Even HCI, as a field, is increasingly aware of the dangers of its own techno-centric visions \cite{SanchezEtAl2025}. Researchers in HRI do repeatedly raise relevant points too \cite{Gamboa2025,Zawieska2024,LemaignanEtAl2026}, so it is a troublesome sign that conference themes seem oblivious to it. As Gamboa \cite{Gamboa2025} illustrates, opening statements that reinforce such unasked-for techno-solutionism are also problematic, but conference themes legitimise them by repeating this framing.

\subsection{Necessary technical progress is deprioritised}

Somewhat ironically, even though claiming that social robots are inevitable and desirable is techno-solutionist, it can nonetheless stifle actual technical development in the field, to the degree that these robots may not actually come to exist. Because the eventual existence of the robot is not in question, it is legitimate to focus on studying the effect their various behaviours or designs might have on users. This helps to determine the exact specification of the robot that will eventually be built. In this, we take for granted that, once the design, behaviour, functionality, and so on is clear, the technical development to deliver these is a given. Studies that evaluate robot design are hence prioritised over technical development needed to deliver the functionality such a social robot would need to have. User studies thus form, for example by far the largest submission category at HRI (Belpaeme, personal communication). 

Technical development is not just about hardware, but, perhaps most importantly, about the skills that a robot would need to be a social agent. As a quick example, the field considers Theory of Mind (ToM) to be one of the most crucial requirements for a social robot, and the False Belief Task in particular makes a regular appearance \cite{ElmadagliRenoux2026}. Elmadagli and Renoux \cite{ElmadagliRenoux2026} criticise this focus altogether, arguing that the field needs to move away from ToM as a relevant construct (which, incidentally is also a more general debate in philosophy and psychology, see \emph{e.g.} \cite{gallagherPracticeMindTheory2001}). Without entering the debate about appropriateness, if the field considers ToM so important, why are there no ToM-capable robots? Scassellati wrote his thesis in 2001 \cite{Scassellati2001}, what  progress has the quarter century since then seen? While there certainly also are papers that attempt implementations, the field seems more busy confirming the possible \emph{effect} of a ToM-able robot than asking how to build one, or, to put it in a way that highlights the contribution the field could be making beyond social robot, \emph{explaining} the underlying mechanism. Chasing effects rather than explanations is a pitfall that has also been criticised in psychology \cite{Cummins2010}. In HRI, the false belief task, implemented using scripted robots, and sometimes even videos thereof \cite{SturgeonEtAl2019,RuoccoEtAl2021} is a good example of both deprioritising technical development and effect-chasing. In psychology, a field that can likewise over-emphasise empirical work, researchers argue for a move away from such an emphasis to focus more on explanations of cognitive capacities \cite{vanRooijBaggio2021}. These capacities refer to the abilities necessary for humans to act in the real world, which would be of obvious interest in HRI too; yet with a frame encouraging effect-chasing instead, we are distracted from such questions.

In a sense, the situation is reminiscent of AI approaches in early robotics, which focussed on the formal reasoning systems necessary to solve complex tasks, while understanding how to tie this to perception and action in the real world was considered secondary. Brooks \cite{Brooks1990} famously rejected this premisse entirely in favour of reactive robots, for which interaction with the world came first. Since then, robotics has shown the value of both. Stanley, the autonomous vehicle that won the DARPA challenge, used a three layer architecture that combined deliberative layers with Brooks' notion of reactive ones \cite{ThrunEtAl2006}. In other words, the point here is not to argue that the field should now go all-in on technical development and abandon user studies; just that framing a social robot as inevitable risks overly deprioritising necessary work that demonstrably has significant potential for HRI.

\subsection{User studies are underconstrained}

Not only is the emphasis in the field on user studies; given that the exact nature of future society remains, obviously, unclear \cite{Gabriel2013}, there is no principled constraint on what constitutes a reasonable scenario to explore in such studies. Any imagined scenario \emph{might} be a possible future scenario, which often seems justification enough to run the study. The need to ground such empirical work in theory is reduced if not eliminated. This risks leading to research that is ``more self-serving than contributing to desirable progress'' \cite[p.219]{Gamboa2025}, and it thus comes as no surprise that the lack of theory in such work has been flagged as a concern before \cite{LeichtmannEtAl2022}. User studies have become a way to produce bricks without checking whether they are needed in those quantities, or even the right shape \cite{Forscher1963}. Recent efforts \cite{GuerraEtAl2026} and proposals \cite{TiltonReig2026} to systematically inventorise tasks and scenarios are therefore much needed, not just for replicability, but also to  better understand why user studies investigate what they do. 

Even if the scenario is plausible, it does not mean that the research done necessarily has an impact on how it plays out in the real world. Waiter robots are a good example here. On the one hand, they are a popular subject in studies and surveys on a range of topics such as user preferences, personalisation, and behaviours \cite{KnightEtAl2024,KimEtAl2021,RossiEtAl2022}. On the other, the waiter robots we encounter in the real world appear to have taken very little from such studies. To the degree that there even is an interaction with the robot, it is reduced to trying to get the food off the tray before the robot moves on to the next task. In the real world, these robots seem more akin to washing machines than to social agents, illustrating that just because a robot could be a social agent, this does not mean it has to be.

\subsection{Wizards no longer disappear}

For all the legitimate uses of the Wizard of Oz (WoZ) paradigm, we also have to recognise its role in both enabling underconstrained user studies and deprioritising technical work because neither would be possible without it. With WoZ, humans can cover for aspects that are difficult if not impossible to implement, such as ToM, and any scenario can be role-played. Herein too lies an irony, because this goes against the original purpose of the paradigm \cite{kelley_empirical_1983,Kelley1984}: a human was to be a stand-in the most difficult aspects that a system under development eventually needed to be capable of (natural language processing in the original work). From initial interactions with this overall system, the developer could gather data to help iteratively replace the wizard with bits of code. Next to data collections during development, the method thus also provides its own validation mechanisms because the code that replaces part of what the human wizard did should not deteriorate the overall performance of the system. Critically, the process is not complete until the wizard has successfully been written out of the system. 

It is very telling that this original purpose of the paradigm is largely lost in HRI work. One exception that comes to mind is the suggestion of Tom Williams and colleagues to implement ``Scarecrows'' \cite{williams_scarecrows_2024}: LLMs can be used as a stand in for actual functionality when it is not yet clear how to develop that functionality, with the explicit purpose of eventually replacing these stand-in functions. In this sense, there is work that makes use of the WoZ paradigm to drive technical development. At the same time, one has to wonder what it says about the field that we now rediscover, apparently without even realising, paradigms that we already had in the 80s.

Overall, the field currently seems framed in a way that deprioritises technical development. That does not mean that none exist; as other venues demonstrate, there is significant progress in hardware development, for example; it just takes place outside of HRI itself. The next section explores why this too is problematic.

\subsection{Robots ``only'' do Kung Fu and fold towels}

One of the most salient hardware developments in robotics at the moment is in robot control, where very agile robots from a number of manufacturers are often (at the time of writing) demonstrated carrying out very fast martial-arts style movements. Technically this is clearly impressive since it demonstrates significant progress not just in the processing necessary but also the quality of the actuators that enable the rapid response times of these robots. However, none of these robots appear to possess any features specifically for \emph{interaction} with humans. Kung Fu robots remain a tech demo that takes place in carefully controlled environments, far removed from scenarios we consider in HRI. Further, as Lemaignan and colleagues have recently discussed, the development of such machines comes with little to no consideration for concepts that we ought to value when we want to build humanoids responsibly \cite{LemaignanEtAl2026} and it plays into the previously discussed techno-solutionism that we should question \cite{SanchezEtAl2025}. By deprioritising technical work in favour of user studies, our conferences themes also put hardware development in the hands of companies with their own agendas. Matheus and Scassellati rightly argue that we need to take robot development back into our own hands \cite{MatheusScassellati2026} and others are right to point out that we need to resist technosolutionist hype in favour of our asserting own values \cite{Gamboa2025,LemaignanEtAl2026}.

That said, we should also not summarily dismiss the Kung Fu robots. There is a nice body of research by Alessandra Sciutti's lab in particular, demonstrating that humans are very perceptive of the specifics of the kinematics of movements \cite{LastricoEtAl2023,VannucciEtAl2024,DiCesareEtAl2020}, so much so that their behaviour changes depending on whether or not a robot's movements are life-like \cite{BisioEtAl2014}. Sciutti and colleagues can demonstrate such effects on a very expensive iCub because the quality of the motors allows such a detailed manipulation. At the same time, the iCub was originally designed for developmental robotics and is not a particularly compelling HRI platform in the sense that very little about it captures what a future social robot might look like, if it will exist, least of all the price tag. The iCub was and remains intended as a research platform only. What the Kung Fu robots show is that it is now possible to build robots that allow this kind of very precise control over the kinematics at a much more accessible price point. If humans and robots do end up interacting in a meaningful way (in the sense of what HRI research assumes), then we can expect, based on research such as that just discussed, that these kind of motors will play an important role in shaping the interaction, even if only in subtle ways. Kung Fu and towel folding are not HRI tasks, and present-day commercial humanoids are not designed for HRI either, but the technological development can be relevant if HRI, as a field, lets it be. This too demonstrates the need to once more focus on the development of robots \cite{MatheusScassellati2026} but for this to happen, we have to stop framing the field in a way that depriorities technological progress. 

As a second point, it is interesting to observe that there is no significant industry effort to build social robots. Companies that tried invariably failed to establish themselves \cite{MatheusScassellati2026} even though some did target compelling markets. Rethink Robotics, for example, marketed Baxter as a collaborative robot for industrial settings  while Softbank Robotics (n\'{e}e Aldebaran) simply targeted researchers interested in social robots rather than applications thereof. When Matheus and Scassellati \cite{MatheusScassellati2026} suggest deployable research products that bridge the gap between lab and product, this also acknowledges that HRI remains first and foremost (and perhaps exclusively) a research field \cite{Zawieska2024}. In this sense too, conference themes that assume robots will eventually be in society seem misplaced: not only do they take bring a techno-solutionist perspective to a research field, robot companies themselves have invariably failed to establish a market here.

With this, we conclude the discussion of the challenges around simply assuming that social robots are inevitable. What is left to discuss is how the field might want to move forward. To that end, we next discuss the need to be more explicit in our work regarding the degree to which it depends on specific visions of the future. We then conclude with a suggestion on how to frame the field to put a renewed focus on the aspects that have become deprioritised. 

\section{A simple kind of dependency test}

Given that the vision of an inevitable social robot has the potential to explicitly or implicitly shape the field in undesirable ways, it appears important to better understand to what degree the relevance of a paper in HRI depends on a future human-robot society becoming reality. If the latter never happens, does the paper have any contribution at all? 

There clearly are user studies that do; for example those that deal with robots that actually do exist (such as waiter robots, delivery robots, and so on), in scenarios that do play out in the real world. Such research can even take, for example, take informative ethnographic forms, see for example \cite{PelikanEtAl2024}). There are also relevant studies that are not, strictly speaking, about robots per se. For example,  humans can attribute intentionality and agency even where there is none \cite{Ziemke2023}, and, robots allow us to study such human cognitive mechanisms even if they will never play a societal role. 

Technical contributions tend to satisfy this test too. When it comes to cognitive abilities of a robot, they force exploring the mechanisms rather than just measuring the effects. If guided by the right theory \cite{Guest2024}, this allows exploring the implications of ideas \cite{McClelland2009}, and can help refine theory \cite{GuestMartin2021a}. Such advances do not depend a human-robot collaborative future. The two papers that were awarded the Test of Time Award by the HRI conference in 2026 similarly have no such dependence: one has developed metrics that are useful when studying human interaction with intelligent technology and sharing findings with each other \cite{SteinfeldEtAl2006}, the other gives insights into how humans interpret and understand motions, and thus how to make sure the intent of a motion is legible \cite{DraganEtAl2013}. 

If nothing else, explicitly asking to what degree a paper depends on a specific vision of the future is thus already helpful in avoiding some of the pitfalls above. The final point to make is that choosing different frames can also guide the field in more fruitful directions.

\section{I is for Interaction}

We have, so far, avoided an explicit definition of what a social robot actual is, essentially because we have been focussed on the frame set by conference themes and paper introductions. To think about alternative frames, however, it is helpful to begin by asking what makes a social robot in HRI stand out from other technology that humans interact with. In HRI, we think of social robots not merely as a something that automates a task (such as laundry folding or dishwasher loading), but as a social agent in some meaningful sense \cite{ZoncaEtAl2021}. One can, of course, study mobile food trays in restaurants, delivery robots, towel folding robots, and the like as part of HRI research, even if the task is not truly HRI, simply because there always is some interaction. However, this remains more in the remit of User Experience Design (UXD) than HRI. Insights from human cognition are clearly also relevant here: Don Normann's ``The design of everyday things'' \cite{Norman2002} relies heavily on human affordance perception, for example. But this does not study what makes a robot social, just how humans interact with technology. Janet Vertesi, who gave a keynote at HRI in 2019, describes an ethnographic study of the team in charge of the Mars Rover \cite{Vertesi2015}. A nice lesson from this is that there is always an interaction with a robot, even when it is in deep space. At the same time, the most important insights from this work arguably come from understanding how the mission contributes to, and indeed shapes, the order and dynamics of the work in the laboratory \cite{Vertesi2012a} and not from considering the rover as a social agent. HRI, however needs to focus on what makes robots social agents, because the answer to that is what differentiates robots in HRI from another technology in UXD. Differently put, there is more to HRI than UXD \cite{Zawieska2024}, and it is in the ways in which HRI goes beyond UXD that we can find new opportunities for the field. 

Focussing on abilities needed for social robots obviously still does not demonstrate the the world actually needs the latter. The future might still be simple automation, and any future robot just an everyday thing \cite{Norman2002}, not an agent. At the same time, successful HRI requires a thorough understanding of those aspects of cognition that are most characteristic of human cognition: the abilities we possess to \emph{interact} with each other in complex social situations. Hanne de Jaegher and colleagues made a very compelling case in 2010 that social interaction itself is \emph{constitutive} of cognition \cite{DeJaegherEtAl2010}, rather than just the outcome of a process inside the heads of individuals. While their criticisms were aimed at the field of social cognition, the insight matters in HRI too. It is the field that produces agents with which we can push the study of such abilities forward. If de Jaegher and colleagues are right (and there is no reason to suspect that they are not), then studying how humans interact with other agents, including artificial ones, does have societal relevance even when social robots themselves do not.

Zawieska \cite{Zawieska2024} concludes with suggestions to extend the concept of ``interaction'' to include integration into everyday lives, and to lived experiences. Here, we have likewise found a way towards realising that the key letter in HRI is the letter ``I''. In a sense, however, Zawieska still presupposes a future with social robots, while we now remain agnostic about such a future. If HRI focusses on what would make a social robot social, then, even when viewed from a future in which social robots never became more than ``a dusty 1980s vision'' \cite{GronbaekEtAl2026}, the field will be seen to have contributed to the understanding of interaction, including, and maybe most importantly, social interaction as a human cognitive mechanism. McClelland shows how the value of modelling in the cognitive sciences lies in the exploration of the implication of ideas \cite{McClelland2009}. Robots conceived in HRI research, to function, need controllers based on our ideas of how humans interact with each other. In that sense, the ``I'' may stand for interaction, but what is studied and understood is how this pertains to the ``H'', while the ``R'' is the means to do so. This also helps us identify the relevant theories in the cognitive sciences \cite{SandiniEtAl2025} that can ground empirical work (as part of which user studies remain relevant too) and progress. None of this requires robots to become societally relevant, but if they do, then this ensures that the deployable research products of the future \cite{MatheusScassellati2026} are grounded in sound fundamental research that is not just begging the question. It is this kind of scientific basis for what is and remains a research field \cite{Zawieska2024} that we lose when our conference themes claim that the social robot is already inevitable.

\section{Conclusion}

To reiterate and summarise this paper: we need to stop pretending social robots are inevitable, whether in opening lines of papers or conference themes. Continuing to operate in such a frame promotes research that does not advance the field at the expense of research that does. It gives license to imagine arbitrary scenarios in user studies while simultaneously putting the emphasis on studying how humans behave in these scenarios rather than the development of the abilities that social robots require. Technological work necessary to build a social agent is deprioritised. Development is instead dominated by techno-solutionist task automation because moving technical development out of the focus of HRI leaves the choice of what robots should be built to companies that neither share our community values \cite{LemaignanEtAl2026} nor even build robots that are relevant for HRI \cite{MatheusScassellati2026}. Robotic Kung Fu or impressive half-marathon times may demonstrate motors and actuation that could well be beneficial for a social robot, but not actual social agents. Tech demos show the industry's best guess at a viable product, such a a towel-folding humanoid, but not what makes social agents social. In the absence of any robot from HRI however, they do shape public (and researcher's) perception, and, by extension, the research field. 

It may well be that task automation is all the world really needs from robots, but HRI as a research field has values and aims that go beyond such products, and we should not lose these out of sight. A frame that never questions if it is even possible or desirable to build a social robot, because it simply declares that it already is inevitable, does not help shape the science of the field. Worse, it risks suffocating the research that is needed for the field to thrive. This is not restricted to research that helps us understand how to actually build social robots. With the social robot inevitable, it is already deemed ``the solution'' to whatever societal issues exist, and we also forego meaningful participatory research on what society really needs. The field should have an interest in resiting such a techno-solutionist positioning \cite{Gamboa2025}. If it were to instead focus on, for example, on technical development grounded in theories about human cognition (e.g. \cite{BisioEtAl2014}), the field would put itself in a position to contribute no matter the future of social robots. We discussed a kind of dependency test earlier: assessing to what degree research makes a contribution to the field even if the particular vision it has of the future does not materialise. This is likely not sufficient as a condition for advancing the field, but it seems necessary.

This paper emphasised the importance of the ``I'' in HRI, showing that a focus on the necessary mechanisms for interaction not only gives access to theoretical grounding for work in HRI that highlights necessary technical advances, it also demonstrates that meaningful work need not be dependent on a social robot actually ever becoming a reality. That said, this is not meant to prescribe a specific direction, just to give an example of one that appears more desirable. There are likely many other frames that have a positive impact on the field. The point is that how we frame the field shapes the research that is carried out. Herein lies a subtle but very important difference between visions of \emph{possible} and of \emph{inevitable} futures: the former intrinsically retains the possibility that the future may not come to pass and frames the research accordingly. Our current conference themes, however, by pushing inevitability rather than possibility, do not set a helpful frame. When we combine this with current industry trends that create hype around humanoid robots which, while technically impressive, are in no way a social agent, we do risk losing what it actually is that sets HRI specifically apart from other fields that study how humans interact with machines that automate tasks. 

To conclude, HRI is and remains a research field \cite{Zawieska2024}. It is therefore also important that it remains in the hands of researchers \cite{LemaignanEtAl2026,MatheusScassellati2026}. It should be driven by fundamental scientific curiosity, not techno-solutionist visions of future robots. The conferences in our field have an active role to play in this, and, by choosing their themes wisely, an easy means to do so. 

It is said that the Germans have a word for everything. In the age of techno-solutionism and hype \cite{Gamboa2025,birhaneValuesEncodedMachine2022,SanchezEtAl2025}, the HRI community needs to find the \emph{Besonnenheit} to not let itself be distracted by either, and to remain committed to the scientific core of the field. The themes of future conferences will demonstrate whether it is able to do so.

\bibliographystyle{IEEEtran}
\bibliography{refs}

@inproceedings{Gamboa2025,
  title = {Robots Are {{Increasingly}}: {{Imagination Crisis}} in {{Human-Computer Interaction Research}}},
  shorttitle = {Robots Are {{Increasingly}}},
  booktitle = {Proceedings of the Sixth Decennial {{Aarhus}} Conference: {{Computing X Crisis}}},
  author = {Gamboa, Mafalda},
  year = {2025},
  series = {{{AAR}} '25},
  pages = {216--222},
  publisher = {Association for Computing Machinery},
  location = {New York, NY, USA},
  doi = {10.1145/3744169.3744189},
  isbn = {979-8-4007-2003-1}
}

@article{vanRooijBaggio2021,
  title = {Theory {Before} the {Test}: {How} to {Build} {High}-{Verisimilitude} {Explanatory} {Theories} in {Psychological} {Science}},
  volume = {16},
  issn = {1745-6916},
  shorttitle = {Theory {Before} the {Test}},
  doi = {10.1177/1745691620970604},
  language = {en},
  number = {4},
  urldate = {2024-02-12},
  journal = {Perspectives on Psychological Science},
  publisher = {SAGE Publications Inc},
  author = {van Rooij, Iris and Baggio, Giosuè},
  month = jul,
  year = {2021},
  pages = {682--697},
}

@article{ThrunEtAl2006,
  title = {Stanley: {The} robot that won the {DARPA} {Grand} {Challenge}},
  volume = {23},
  copyright = {Copyright © 2006 Wiley Periodicals, Inc., A Wiley Company},
  issn = {1556-4967},
  shorttitle = {Stanley},
  doi = {10.1002/rob.20147},
  language = {en},
  number = {9},
  urldate = {2026-06-12},
  journal = {Journal of Field Robotics},
  author = {Thrun, Sebastian and Montemerlo, Mike and Dahlkamp, Hendrik and Stavens, David and Aron, Andrei and Diebel, James and Fong, Philip and Gale, John and Halpenny, Morgan and Hoffmann, Gabriel and Lau, Kenny and Oakley, Celia and Palatucci, Mark and Pratt, Vaughan and Stang, Pascal and Strohband, Sven and Dupont, Cedric and Jendrossek, Lars-Erik and Koelen, Christian and Markey, Charles and Rummel, Carlo and van Niekerk, Joe and Jensen, Eric and Alessandrini, Philippe and Bradski, Gary and Davies, Bob and Ettinger, Scott and Kaehler, Adrian and Nefian, Ara and Mahoney, Pamela},
  year = {2006},
  note = {\_eprint: https://onlinelibrary.wiley.com/doi/pdf/10.1002/rob.20147},
  pages = {661--692},
}

@article{Brooks1990,
  series = {Designing {Autonomous} {Agents}},
  title = {Elephants don't play chess},
  volume = {6},
  issn = {0921-8890},
  url = {https://www.sciencedirect.com/science/article/pii/S0921889005800259},
  doi = {10.1016/S0921-8890(05)80025-9},
  number = {1},
  urldate = {2025-12-05},
  journal = {Robotics and Autonomous Systems},
  author = {Brooks, Rodney A.},
  month = jun,
  year = {1990},
  pages = {3--15},
}

@phdthesis{Scassellati2001,
  title = {Foundations for a theory of mind for a humanoid robot},
  language = {eng},
  urldate = {2026-06-12},
  school = {Massachusetts Institute of Technology},
  author = {Scassellati, Brian M.},
  year = {2001},
}

@article{gallagherPracticeMindTheory2001,
  title = {The {Practice} of {Mind}: {Theory}, {Simulation} or {Primary} {Interaction}?},
  volume = {8},
  shorttitle = {The {Practice} of {Mind}},
  number = {5-7},
  journal = {Journal of Consciousness Studies},
  publisher = {Imprint Academic},
  author = {Gallagher, Shaun},
  year = {2001},
  pages = {83--108},
}

@inproceedings{ElmadagliRenoux2026,
  address = {New York, NY, USA},
  series = {{HRI} '26},
  title = {Theory of {Whose} {Mind}? {Exposing} the {Shortcomings} of {One} of {HRI}’s {Core} {Concepts}},
  isbn = {979-8-4007-2128-1},
  shorttitle = {Theory of {Whose} {Mind}?},

  doi = {10.1145/3757279.3788818},
  urldate = {2026-06-11},
  booktitle = {Proceedings of the 21st {ACM}/{IEEE} {International} {Conference} on {Human}-{Robot} {Interaction}},
  publisher = {Association for Computing Machinery},
  author = {Elmadagli, Cansu and Renoux, Jennifer},
  month = mar,
  year = {2026},
  pages = {1369--1377},
}

@inproceedings{RuoccoEtAl2021,
  address = {New York, NY, USA},
  series = {{HAI} '21},
  title = {Theory of {Mind} {Improves} {Human}’s {Trust} in an {Iterative} {Human}-{Robot} {Game}},
  isbn = {978-1-4503-8620-3},
  doi = {10.1145/3472307.3484176},
  urldate = {2026-06-11},
  booktitle = {Proceedings of the 9th {International} {Conference} on {Human}-{Agent} {Interaction}},
  publisher = {Association for Computing Machinery},
  author = {Ruocco, Martina and Mou, Wenxuan and Cangelosi, Angelo and Jay, Caroline and Zanatto, Debora},
  month = nov,
  year = {2021},
  pages = {227--234},
}

@inproceedings{SturgeonEtAl2019,
  title = {Perception of {Social} {Intelligence} in {Robots} {Performing} {False}-{Belief} {Tasks}},
  issn = {1944-9437},

  doi = {10.1109/RO-MAN46459.2019.8956467},
  urldate = {2026-06-12},
  booktitle = {2019 28th {IEEE} {International} {Conference} on {Robot} and {Human} {Interactive} {Communication} ({RO}-{MAN})},
  author = {Sturgeon, Stephanie and Palmer, Andrew and Blankenburg, Janelle and Feil-Seifer, David},
  month = oct,
  year = {2019},
  note = {ISSN: 1944-9437},
  pages = {1--7},
}

@incollection{Cummins2010,
  title = {How Does It {{Work}}?' Vs. `{{What}} Are the {{Laws}}?},
  booktitle = {The {{World}} in the {{Head}}},
  author = {Cummins, Robert},
  editor = {Cummins, Robert},
  year = 2010,
  pages = {282--310},
  publisher = {Oxford University Press}
}

@article{BartlettEdmundsBelpaeme2022,
  title = {Have {{I}} Got the Power? {{Analysing}} and Reporting Statistical Power in {{HRI}}},
  author = {Bartlett, Madeleine E. and Edmunds, C. E. R. and Belpaeme, Tony and Thill, Serge},
  year = 2022,
  journal = {ACM Transactions on Human-Robot Interaction},
  volume = {11},
  number = {2},
  doi = {10.1145/3495246}
}

@article{GuerraEtAl2026,
  title = {Understanding {{Task Diversity}} in {{Human-Robot Collaboration}}: {{A Scoping Review}}},
  shorttitle = {Understanding {{Task Diversity}} in {{Human-Robot Collaboration}}},
  author = {Guerra, Enrico and B{\"u}ttner, Sebastian Thomas and Prilla, Michael},
  year = 2026,
  month = jan,
  journal = {International Journal of Human--Computer Interaction},
  volume = {0},
  number = {0},
  pages = {1--21},
  publisher = {Taylor \& Francis},
  issn = {1044-7318},
  doi = {10.1080/10447318.2025.2606215}
}

@inproceedings{TiltonReig2026,
  title = {Building a {{Task Repository}} for {{Human-Robot Interaction Studies}}: {{An Exploratory Review Approach}}},
  shorttitle = {Building a {{Task Repository}} for {{Human-Robot Interaction Studies}}},
  booktitle = {Companion {{Proceedings}} of the 21st {{ACM}}/{{IEEE International Conference}} on {{Human-Robot Interaction}}},
  author = {Tilton, Dylan and Reig, Samantha},
  year = 2026,
  month = mar,
  series = {{{HRI Companion}} '26},
  pages = {614--618},
  publisher = {Association for Computing Machinery},
  address = {New York, NY, USA},
  doi = {10.1145/3776734.3794468},
  isbn = {979-8-4007-2321-6}
}

@article{LeichtmannEtAl2022,
  title = {Crisis {{Ahead}}? {{Why Human-Robot Interaction User Studies May Have Replicability Problems}} and {{Directions}} for {{Improvement}}},
  shorttitle = {Crisis {{Ahead}}?},
  author = {Leichtmann, Benedikt and Nitsch, Verena and Mara, Martina},
  year = 2022,
  month = mar,
  journal = {Frontiers in Robotics and AI},
  volume = {9},
  publisher = {Frontiers},
  issn = {2296-9144},
  doi = {10.3389/frobt.2022.838116},
  langid = {english}
}

@article{BethelMurphy2010,
  title = {Review of {{Human Studies Methods}} in {{HRI}} and {{Recommendations}}},
  author = {Bethel, Cindy L. and Murphy, Robin R.},
  year = 2010,
  month = dec,
  journal = {International Journal of Social Robotics},
  volume = {2},
  number = {4},
  pages = {347--359},
  issn = {1875-4805},
  doi = {10.1007/s12369-010-0064-9},
  langid = {english}
}

@article{FelinEtAl2021,
  title = {The Data-Hypothesis Relationship},
  author = {Felin, Teppo and Koenderink, Jan and Krueger, Joachim I. and Noble, Denis and Ellis, George F.R.},
  year = 2021,
  month = feb,
  journal = {Genome Biology},
  volume = {22},
  number = {1},
  pages = {57},
  issn = {1474-760X},
  doi = {10.1186/s13059-021-02276-4},
  langid = {english}
}

@article{BigmanEtAl2026,
  title = {Human–{{AI}} Interaction Research Needs to Be Embedded in Psychological Theory},
  author = {Bigman, Yochanan E. and Briker, Roman and Langer, Markus},
  year = {2026-03-20},
  journal = {Nature Reviews Psychology},
  shortjournal = {Nat Rev Psychol},
  pages = {1--2},
  publisher = {Nature Publishing Group},
  issn = {2731-0574},
  doi = {10.1038/s44159-026-00551-4},
  langid = {english}
}

@inproceedings{DraganEtAl2013,
  title = {Legibility and Predictability of Robot Motion},
  booktitle = {Proceedings of the 8th {{ACM}}/{{IEEE}} International Conference on {{Human-robot}} Interaction},
  author = {Dragan, Anca D. and Lee, Kenton C.T. and Srinivasa, Siddhartha S.},
  year = {2013},
  series = {{{HRI}} '13},
  pages = {301--308},
  publisher = {IEEE Press},
  location = {Tokyo, Japan},
  isbn = {978-1-4673-3055-8}
}

@article{MahonCaramazza2008,
	series = {Links and {Interactions} {Between} {Language} and {Motor} {Systems} in the {Brain}},
	title = {A critical look at the embodied cognition hypothesis and a new proposal for grounding conceptual content},
	volume = {102},
	issn = {0928-4257},
	doi = {10.1016/j.jphysparis.2008.03.004},
	number = {1},
	journal = {Journal of Physiology-Paris},
	author = {Mahon, Bradford Z. and Caramazza, Alfonso},
	month = jan,
	year = {2008},
	pages = {59--70},
}

@article{SandiniEtAl2025,
  title = {Mutual Human-Robot Understanding for a Robot-Enhanced Society: The Crucial Development of Shared Embodied Cognition},
  shorttitle = {Mutual Human-Robot Understanding for a Robot-Enhanced Society},
  author = {Sandini, Giulio and Sciutti, Alessandra and Morasso, Pietro},
  year = {2025},
  journal = {Frontiers in Artificial Intelligence},
  shortjournal = {Front. Artif. Intell.},
  volume = {8},
  publisher = {Frontiers},
  issn = {2624-8212},
  doi = {10.3389/frai.2025.1608014},
  langid = {english}
}

@inproceedings{SteinfeldEtAl2006,
  title = {Common Metrics for Human-Robot Interaction},
  booktitle = {Proceedings of the 1st {{ACM SIGCHI}}/{{SIGART}} Conference on {{Human-robot}} Interaction},
  author = {Steinfeld, Aaron and Fong, Terrence and Kaber, David and Lewis, Michael and Scholtz, Jean and Schultz, Alan and Goodrich, Michael},
  year = {2006},
  series = {{{HRI}} '06},
  pages = {33--40},
  publisher = {Association for Computing Machinery},
  location = {New York, NY, USA},
  doi = {10.1145/1121241.1121249},
  isbn = {978-1-59593-294-5}
}

@inproceedings{LemaignanEtAl2026,
  title = {Responsible {{Humanoids}}: {{A Contradiction}} in {{Terms}}?},
  shorttitle = {Responsible {{Humanoids}}},
  booktitle = {Proceedings of the 21st {{ACM}}/{{IEEE International Conference}} on {{Human-Robot Interaction}}},
  author = {Lemaignan, Séverin and Moon, Ajung and Coghlan, Simon and Collins, Emily C. and Evers, Vanessa and Hochgeschwender, Nico and Ljungblad, Sara and Milford, Michael and Moth-Lund Christensen, Sarah and Rodríguez Lera, Francisco J. and Salvini, Pericle and Yang, Yi},
  year = {2026},
  series = {{{HRI}} '26},
  pages = {1341--1349},
  publisher = {Association for Computing Machinery},
  location = {New York, NY, USA},
  doi = {10.1145/3757279.3788817},
  isbn = {979-8-4007-2128-1}	
}

@inproceedings{MatheusScassellati2026,
	address = {New York, NY, USA},
	series = {{HRI} '26},
	title = {We {Cannot} {Outsource} {What} {We} {Value} {Most}: {Toward} {Deployable} {Research} {Products} in {HRI}},
	isbn = {979-8-4007-2128-1},
	shorttitle = {We {Cannot} {Outsource} {What} {We} {Value} {Most}},
	doi = {10.1145/3757279.3788816},
	booktitle = {Proceedings of the 21st {ACM}/{IEEE} {International} {Conference} on {Human}-{Robot} {Interaction}},
	publisher = {Association for Computing Machinery},
	author = {Matheus, Kayla and Scassellati, Brian},
	month = mar,
	year = {2026},
	pages = {1359--1368},
}

@inproceedings{LastricoEtAl2023,
	title = {Expressing and {Inferring} {Action} {Carefulness} in {Human}-to-{Robot} {Handovers}},
	issn = {2153-0866},
	doi = {10.1109/IROS55552.2023.10342111},
	booktitle = {2023 {IEEE}/{RSJ} {International} {Conference} on {Intelligent} {Robots} and {Systems} ({IROS})},
	author = {Lastrico, Linda and Duarte, Nuno Ferreira and Carfí, Alessandro and Rea, Francesco and Sciutti, Alessandra and Mastrogiovanni, Fulvio and Santos-Victor, José},
	month = oct,
	year = {2023},
	note = {ISSN: 2153-0866},
	pages = {9824--9831},
}

@article{VannucciEtAl2024,
	title = {Humanoid {Attitudes} {Influence} {Humans} in {Video} and {Live} {Interactions}},
	volume = {12},
	issn = {2169-3536},
	doi = {10.1109/ACCESS.2024.3442863},
	journal = {IEEE Access},
	author = {Vannucci, Fabio and Lombardi, Giada and Rea, Francesco and Sandini, Giulio and Di Cesare, Giuseppe and Sciutti, Alessandra},
	year = {2024},
	pages = {118502--118509},
}

@article{DiCesareEtAl2020,
	title = {How attitudes generated by humanoid robots shape human brain activity},
	volume = {10},
	copyright = {2020 The Author(s)},
	issn = {2045-2322},
	doi = {10.1038/s41598-020-73728-3},
	language = {en},
	number = {1},
	journal = {Scientific Reports},
	publisher = {Nature Publishing Group},
	author = {Di Cesare, G. and Vannucci, F. and Rea, F. and Sciutti, A. and Sandini, G.},
	month = oct,
	year = {2020},
	pages = {16928},
}

@article{Forscher1963,
  title = {Chaos in the {{Brickyard}}},
  author = {Forscher, Bernard K.},
  year = 1963,
  month = oct,
  journal = {Science},
  volume = {142},
  number = {3590},
  pages = {339--339},
  publisher = {American Association for the Advancement of Science},
  doi = {10.1126/science.142.3590.339.a}
}

@article{BisioEtAl2014,
	title = {Motor {Contagion} during {Human}-{Human} and {Human}-{Robot} {Interaction}},
	volume = {9},
	issn = {1932-6203},
	doi = {10.1371/journal.pone.0106172},
	language = {en},
	number = {8},
	journal = {PLOS ONE},
	publisher = {Public Library of Science},
	author = {Bisio, Ambra and Sciutti, Alessandra and Nori, Francesco and Metta, Giorgio and Fadiga, Luciano and Sandini, Giulio and Pozzo, Thierry},
	month = aug,
	year = {2014},
	pages = {e106172},
}

@article{Zawieska2024,
  title = {The {{Iron Cage}} of {{Social Robotics}}},
  author = {Zawieska, Karolina},
  year = 2024,
  month = dec,
  journal = {ACM Transactions on Human-Robot Interaction},
  volume = {14},
  number = {1},
  pages = {18:1--18:10},
  doi = {10.1145/3695772}
}

@article{ZoncaEtAl2021,
  title = {The Role of Reciprocity in Human-Robot Social Influence},
  author = {Zonca, Joshua and Folsø, Anna and Sciutti, Alessandra},
  year = {2021},
  journal = {iScience},
  shortjournal = {iScience},
  volume = {24},
  number = {12},
  pages = {103424},
  issn = {2589-0042},
  doi = {10.1016/j.isci.2021.103424}
}

@article{williams_scarecrows_2024,
	title = {Scarecrows in {Oz}: {The} {Use} of {Large} {Language} {Models} in {HRI}},
	volume = {13},
	shorttitle = {Scarecrows in {Oz}},
	doi = {10.1145/3606261},
	number = {1},
	journal = {J. Hum.-Robot Interact.},
	author = {Williams, Tom and Matuszek, Cynthia and Mead, Ross and Depalma, Nick},
	month = jan,
	year = {2024},
	pages = {1:1--1:11},

}

@article{McClelland2009,
  title = {The {{Place}} of {{Modeling}} in {{Cognitive Science}}},
  author = {McClelland, James L.},
  year = {2009},
  journal = {Topics in Cognitive Science},
  volume = {1},
  number = {1},
  pages = {11--38},
  issn = {1756-8765},
  doi = {10.1111/j.1756-8765.2008.01003.x},
  langid = {english}
}

@inproceedings{birhaneValuesEncodedMachine2022,
  title = {The {{Values Encoded}} in {{Machine Learning Research}}},
  booktitle = {Proceedings of the 2022 {{ACM Conference}} on {{Fairness}}, {{Accountability}}, and {{Transparency}}},
  author = {Birhane, Abeba and Kalluri, Pratyusha and Card, Dallas and Agnew, William and Dotan, Ravit and Bao, Michelle},
  year = {2022},
  series = {{{FAccT}} '22},
  pages = {173--184},
  publisher = {Association for Computing Machinery},
  location = {New York, NY, USA},
  doi = {10.1145/3531146.3533083},
  isbn = {978-1-4503-9352-2}
}

@article{DeJaegherEtAl2010,
  title = {Can Social Interaction Constitute Social Cognition?},
  author = {De Jaegher, Hanne and Di Paolo, Ezequiel and Gallagher, Shaun},
  year = {2010},
  journal = {Trends in Cognitive Sciences},
  shortjournal = {Trends in Cognitive Sciences},
  volume = {14},
  number = {10},
  pages = {441--447},
  issn = {1364-6613},
  doi = {10.1016/j.tics.2010.06.009}
}

@book{Vertesi2015,
  title = {Seeing {{Like}} a {{Rover}}: {{How Robots}}, {{Teams}}, and {{Images Craft Knowledge}} of {{Mars}}},
  shorttitle = {Seeing {{Like}} a {{Rover}}},
  year = {2015},
  author = {Vertesi, Janet},
  publisher = {University of Chicago Press},
  location = {Chicago, IL},
  langid = {english},
  pagetotal = {304}
}

@article{Selwyn2022,
  title = {The Future of {{AI}} and Education: {{Some}} Cautionary Notes},
  shorttitle = {The Future of {{AI}} and Education},
  author = {Selwyn, Neil},
  year = {2022},
  journal = {European Journal of Education},
  volume = {57},
  number = {4},
  pages = {620--631},
  issn = {1465-3435},
  doi = {10.1111/ejed.12532},
  langid = {english}
}

@inproceedings{GronbaekEtAl2026,
  title = {How {{Do Future Visions Shape}} the {{Field}} of {{Human-Computer Interaction}}?},
  booktitle = {Proceedings of the 2026 {{CHI Conference}} on {{Human Factors}} in {{Computing Systems}} ({{CHI}} ’26)},
  author = {Grønbæk, Jens Emil and Klokmose, Clemens Nylandsted and Hornbæk, Kasper},
  year = {2026},
  doi = {10.1145/3772318.3791038},
  langid = {english}
}

@article{Guest2024,
  title = {What {{Makes}} a {{Good Theory}}, and {{How Do We Make}} a {{Theory Good}}?},
  author = {Guest, Olivia},
  year = {2024},
  journal = {Computational Brain \& Behavior},
  shortjournal = {Comput Brain Behav},
  issn = {2522-087X},
  doi = {10.1007/s42113-023-00193-2},
  langid = {english}
}

@inproceedings{RossiEtAl2022,
  title = {Investigating {{Customers}}' {{Preferences}} of {{Robot}}'s {{Serving Styles}}},
  booktitle = {Proceedings of the 2022 {{ACM}}/{{IEEE International Conference}} on {{Human-Robot Interaction}}},
  author = {Rossi, Alessandra and Caputo, Alessandro and Scafora, Alessio and Rossi, Silvia},
  year = {2022},
  series = {{{HRI}} '22},
  pages = {1017--1020},
  publisher = {IEEE Press},
  location = {Sapporo, Hokkaido, Japan}
}

@inproceedings{KimEtAl2021,
  title = {On the {{Common}} and {{Different Expectations}} on {{Robot Service}} in {{Restaurant}} between {{Customers}} and {{Employees}}},
  booktitle = {Companion of the 2021 {{ACM}}/{{IEEE International Conference}} on {{Human-Robot Interaction}}},
  author = {Kim, Min-Gyu and Park, Minjung and Kim, Juhyun and Kwon, Yong-Seoup and Sohn, Dong-Seop and Yoon, Heeyoon and Seo, Kap-Ho},
  year= {2021},
  series = {{{HRI}} '21 {{Companion}}},
  pages = {262--265},
  publisher = {Association for Computing Machinery},
  location = {New York, NY, USA},
  doi = {10.1145/3434074.3447172},
  isbn = {978-1-4503-8290-8}
}

@inproceedings{KnightEtAl2024,
  title = {Iterative {{Robot Waiter Algorithm Design}}: {{Service Expectations}} and {{Social Factors}}},
  shorttitle = {Iterative {{Robot Waiter Algorithm Design}}},
  booktitle = {Proceedings of the 2024 {{ACM}}/{{IEEE International Conference}} on {{Human-Robot Interaction}}},
  author = {Knight, Heather and Flynn, Deanna and Oo, Theing Mwe and Hansen, Julia},
  year = {2024},
  series = {{{HRI}} '24},
  pages = {394--402},
  publisher = {Association for Computing Machinery},
  location = {New York, NY, USA},
  doi = {10.1145/3610977.3634978},
  isbn = {979-8-4007-0322-5}
}

@inproceedings{PelikanEtAl2024,
  title = {Encountering {{Autonomous Robots}} on {{Public Streets}}},
  booktitle = {Proceedings of the 2024 {{ACM}}/{{IEEE International Conference}} on {{Human-Robot Interaction}}},
  author = {Pelikan, Hannah R. M. and Reeves, Stuart and Cantarutti, Marina N.},
  year = {2024},
  series = {{{HRI}} '24},
  pages = {561--571},
  publisher = {Association for Computing Machinery},
  location = {New York, NY, USA},
  doi = {10.1145/3610977.3634936},
  isbn = {979-8-4007-0322-5}
}

@inproceedings{vanderBijl-Brouwer2022,
  title = {Design, One Piece of the Puzzle: {{A}} Conceptual and Practical Perspective on Transdisciplinary Design},
  booktitle = {{{DRS Biennial Conference Series}}},
  author = {van der Bijl-Brouwer, Mieke},
  year = {2022},
  doi = {10.21606/drs.2022.402}
}

@article{GuestMartin2021a,
  title = {How {{Computational Modeling Can Force Theory Building}} in {{Psychological Science}}},
  author = {Guest, Olivia and Martin, Andrea E.},
  year = {2021},
  journal = {Perspectives on Psychological Science},
  shortjournal = {Perspect Psychol Sci},
  volume = {16},
  number = {4},
  pages = {789--802},
  publisher = {SAGE Publications Inc},
  issn = {1745-6916},
  doi = {10.1177/1745691620970585},
  langid = {english}
}

@inproceedings{SanchezEtAl2025,
  title = {Let's {{Talk Futures}}: {{A Literature Review}} of {{HCI}}'s {{Future Orientation}}},
  shorttitle = {Let's {{Talk Futures}}},
  booktitle = {Proceedings of the 2025 {{CHI Conference}} on {{Human Factors}} in {{Computing Systems}}},
  author = {Sanchez, Camilo and Wang, Sui and Savolainen, Kaisa and Epp, Felix Anand and Salovaara, Antti},
  year = {2025},
  series = {{{CHI}} '25},
  pages = {1--36},
  publisher = {Association for Computing Machinery},
  location = {New York, NY, USA},
  doi = {10.1145/3706598.3713759},
  isbn = {979-8-4007-1394-1}
}

@book{Norman2002,
  title = {The {{Design}} of {{Everyday Things}}},
  author = {Norman, Donald A.},
  year = {2002},
  publisher = {Basic Books, Inc.},
  location = {USA},
  isbn = {978-0-465-06710-7},
  pagetotal = {272}
}

@article{Vertesi2012a,
  title = {Seeing like a {{Rover}}: {{Visualization}}, Embodiment, and Interaction on the {{Mars Exploration Rover Mission}}},
  shorttitle = {Seeing like a {{Rover}}},
  author = {Vertesi, Janet},
  year = {2012},
  journal = {Social Studies of Science},
  shortjournal = {Soc Stud Sci},
  volume = {42},
  number = {3},
  pages = {393--414},
  publisher = {SAGE Publications Ltd},
  issn = {0306-3127},
  doi = {10.1177/0306312712444645},
  langid = {english}
}

@article{Kelley1984,
	title = {An iterative design methodology for user-friendly natural language office information applications},
	volume = {2},
	issn = {1046-8188},
	doi = {10.1145/357417.357420},
	number = {1},
	journal = {ACM Trans. Inf. Syst.},
	author = {Kelley, J. F.},
	month = jan,
	year = {1984},
	pages = {26--41},
}

@inproceedings{kelley_empirical_1983,
	address = {New York, NY, USA},
	series = {{CHI} '83},
	title = {An empirical methodology for writing user-friendly natural language computer applications},
	isbn = {978-0-89791-121-4},
	doi = {10.1145/800045.801609},
	booktitle = {Proceedings of the {SIGCHI} {Conference} on {Human} {Factors} in {Computing} {Systems}},
	publisher = {Association for Computing Machinery},
	author = {Kelley, J. F.},
	month = dec,
	year = {1983},
	pages = {193--196},
}

@inproceedings{LeeEtAl2021,
  address = {New York, NY, USA},
  series = {{CHI} '21},
  title = {Conversational {Futures}: {Emancipating} {Conversational} {Interactions} for {Futures} {Worth} {Wanting}},
  isbn = {978-1-4503-8096-6},
  shorttitle = {Conversational {Futures}},
  url = {https://dl.acm.org/doi/10.1145/3411764.3445244},
  doi = {10.1145/3411764.3445244},
  urldate = {2026-06-13},
  booktitle = {Proceedings of the 2021 {CHI} {Conference} on {Human} {Factors} in {Computing} {Systems}},
  publisher = {Association for Computing Machinery},
  author = {Lee, Minha and Noortman, Renee and Zaga, Cristina and Starke, Alain and Huisman, Gijs and Andersen, Kristina},
  month = may,
  year = {2021},
  pages = {1--13},
}

@article{Ziemke2023,
  title = {Understanding {{Social Robots}}: {{Attribution}} of {{Intentional Agency}} to {{Artificial}} and {{Biological Bodies}}},
  shorttitle = {Understanding {{Social Robots}}},
  author = {Ziemke, Tom},
  year = {2023},
  journal = {Artificial Life},
  shortjournal = {Artif Life},
  volume = {29},
  number = {3},
  pages = {351--366},
  issn = {1064-5462},
  doi = {10.1162/artl_a_00404}
}

@incollection{PeetersEtAl,
  title = {Disembodied {{Learning}}: {{A Critical Perspective}} on {{Flourishing}} with {{Ai}} in {{Education}}},
  shorttitle = {Disembodied {{Learning}}},
  booktitle = {Shaping {{Children}}'s {{Learning Through Technology}}},
  author = {Peeters, Anco and Chichirău, Mălina and Coggins, Thomasin N. and Thill, Serge},
  editor = {Costescu, Cristina},
  year = {2026},
  publisher = {Springer},
  url  = {https://philarchive.org/rec/PEEDLA}
}

@article{Gabriel2013,
  title = {A Scientific Enquiry into the Future},
  author = {Gabriel, Johannes},
  year = {2013},
  journal = {European Journal of Futures Research},
  shortjournal = {Eur J Futures Res},
  volume = {2},
  number = {1},
  pages = {31},
  issn = {2195-2248},
  doi = {10.1007/s40309-013-0031-4},
  langid = {english}
}

@article{BorupEtAl2006,
  title = {The Sociology of Expectations in Science and Technology},
  author = {Borup, Mads and Brown, Nik and Konrad, Kornelia and Van Lente, Harro},
  year = {2006},
  journal = {Technology Analysis \& Strategic Management},
  volume = {18},
  number = {3--4},
  pages = {285--298},
  publisher = {Routledge},
  issn = {0953-7325},
  doi = {10.1080/09537320600777002}
}

\end{document}